\documentclass{article}
\usepackage{spconf,amsmath,graphicx}
\usepackage{amsmath}
\usepackage{spconf,amsmath,graphicx}
\usepackage{booktabs,siunitx}
\usepackage{xcolor}

\usepackage{multirow}  
\usepackage{array}     
\usepackage{booktabs}  
\usepackage{array}
\usepackage{graphicx} 
\usepackage{tabularx} 
\usepackage{caption} 
\usepackage{hyperref}
\usepackage{amssymb}

\title{SINGLE IMAGE LDR TO HDR CONVERSION using CONDITIONAL Diffusion}

\name{Dwip Dalal$^{\star}$ \qquad Gautam Vashishtha$^{\star}$ \qquad Prajwal Singh$^{\dagger}$ \thanks{$^{\star}$Equal contribution and $^{\dagger}$This work is supported by Prime Minister Research Fellowship (PMRF-2122-2557) and $^\ddagger$ Jibaben Patel Chair.} \qquad Shanmuganathan Raman$^{\ddagger}$ }

			\address{Computer Vision, Imaging and Graphics Lab \\
			Indian Institute of Technology Gandhinagar, India \\ \{dwip.dalal, gautam.pv, singh\_prajwal,
                shanmuga\}@iitgn.ac.in}
                
\begin{document}
\ninept
\maketitle
\begin{abstract}
Digital imaging aims to replicate realistic scenes, but Low Dynamic Range (LDR) cameras cannot represent the wide dynamic range of real scenes, resulting in under-/overexposed images. This paper presents a deep learning-based approach for recovering intricate details from shadows and highlights while reconstructing High Dynamic Range (HDR) images. We formulate the problem as an image-to-image (I2I) translation task and propose a conditional Denoising Diffusion Probabilistic Model (DDPM) based framework using classifier-free guidance. We incorporate a deep CNN-based autoencoder in our proposed framework to enhance the quality of the latent representation of the input LDR image used for conditioning. Moreover, we introduce a new loss function for LDR-HDR translation tasks, termed Exposure Loss. This loss helps direct gradients in the opposite direction of the saturation, further improving the results' quality. By conducting comprehensive quantitative and qualitative experiments, we have effectively demonstrated the proficiency of our proposed method. The results indicate that a simple conditional diffusion-based method can replace the complex camera pipeline-based architectures.

\end{abstract}

\section{Introduction}
\label{sec:intro}

High dynamic range (HDR) imaging is a promising technique for improving the viewing experience of digital content by capturing real-world lighting and details. However, low dynamic range (LDR) cameras are unable to capture the vast dynamic range in real-world scenes. A workaround for this is to 
capture several LDR images taken at various exposures and concatenate them to get HDR images, but it frequently results in ghosting artifacts, especially in scenes that are dynamic. To overcome these limitations, deep convolutional neural networks (CNNs) have been used to develop single-image HDR reconstruction techniques \cite{liu2020single}, \cite{raipurkar2021hdr}, \cite{eilertsen2017hdr}. These techniques address the problems with LDR-to-HDR mapping, which is difficult because HDR pixels (32-bit floating point) have much more variation than LDR pixels (8-bit unsigned integers).





\begin{table}[!tbp]
\centering

\resizebox{0.49\textwidth}{!}{
\label{tab:mytable}
\begin{tabularx}{\textwidth}{@{}lXXX@{}}
\rotatebox[origin=l]{90}{\textbf{ 
  \Large LDR input images}} & \includegraphics[width=\linewidth]{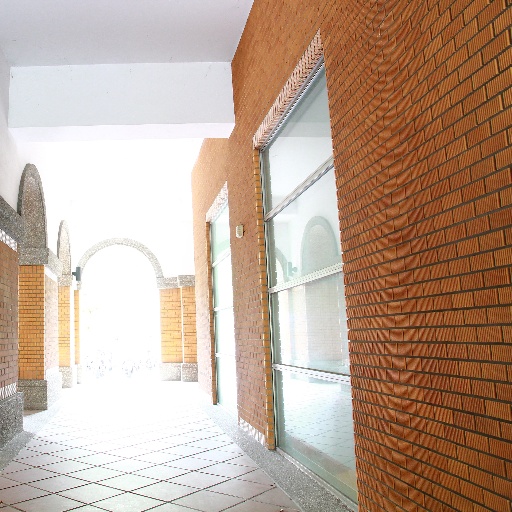} & \includegraphics[width=\linewidth]{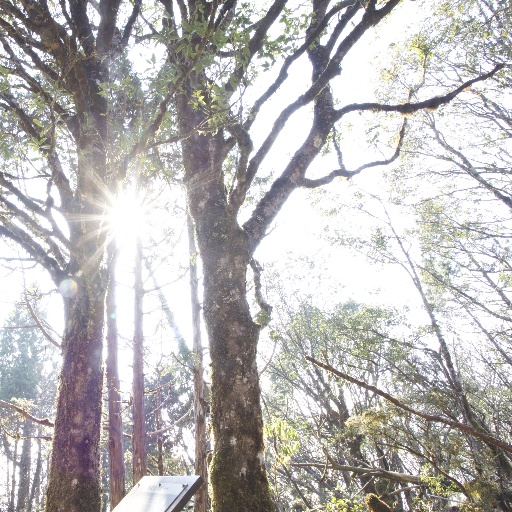} & \includegraphics[width=\linewidth]{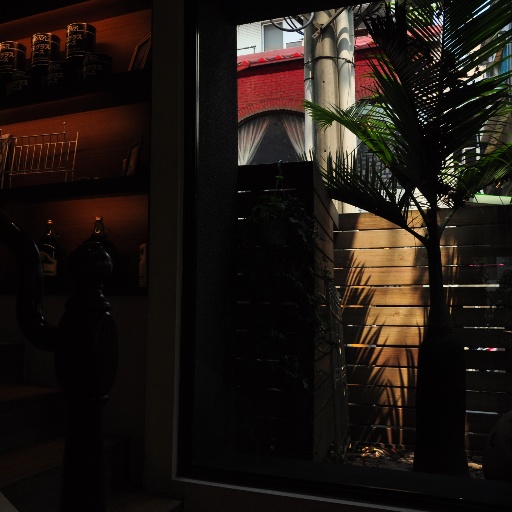} \\
\rotatebox[origin=l]{90}{\textbf{ 
  \Large Our results}} & \includegraphics[width=\linewidth]{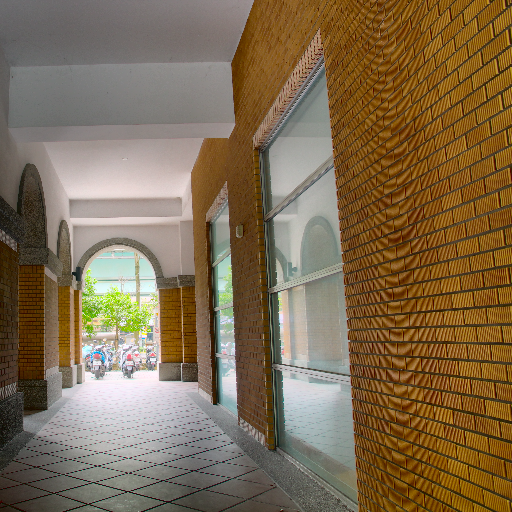} & \includegraphics[width=\linewidth]{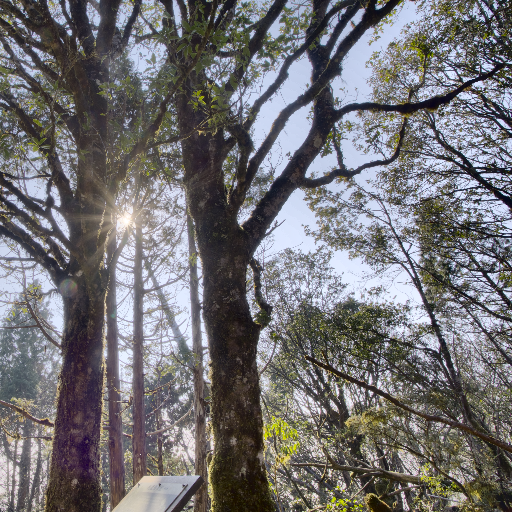} & \includegraphics[width=\linewidth]{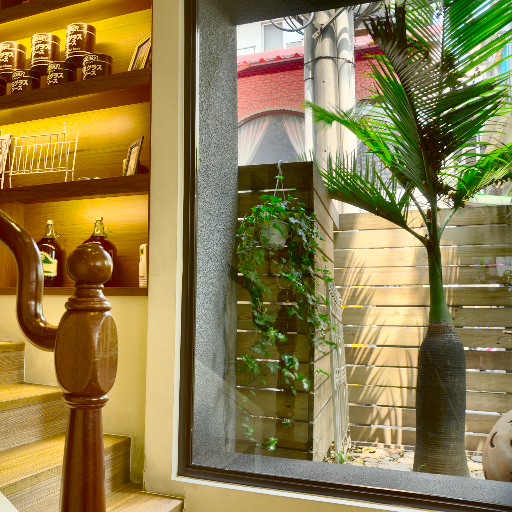} \\
\end{tabularx}
}
\caption{HDR reconstruction from a single LDR image. We condition the DDPM and guide the gradient to recover missing details from shadows and highlights.}
\vspace{-5mm}
\end{table}

Research on LDR to HDR translation has recently received intense attention. Endo et al. \cite{endo2017deep} proposed the deep-learning-based approach for fully automatic inference using deep convolutional neural networks. They adopt a bracketed approach by inferring from a sequence of k LDR images of different exposures and then reconstructing an HDR image by merging the LDR images \cite{debevec2008recovering}. The work \cite{eilertsen2017hdr} used a full CNN design in the form of a hybrid dynamic range autoencoder that transformed the LDR input image using the encoder network to generate a compact feature representation and generated the output HDR image by passing it to the HDR decoder operating in the log domain.   

Liu et al. \cite{liu2020single} approached to tackle this problem by using the domain knowledge of LDR image formation pipeline to decompose the reconstruction into three sub-tasks of i) dequantization, ii) linearization, and iii) hallucination. They developed networks for each of these tasks using CNNs at the core of each architecture. The work in \cite{raipurkar2021hdr} approaches the problem in a similar fashion but utilizes a condition GAN-based framework.


In this paper, we introduce a novel method for reconstructing high-quality HDR images from a single LDR image without the requirement of an explicit inverse camera pipeline \cite{liu2020single} \cite{raipurkar2021hdr}. Our approach  relies upon the utilization of a conditional classifier-free diffusion architecture \cite{ho2022classifier}. Besides the fundamental structure of the diffusion architecture, the model includes an encoder network that generates a latent representation of the LDR input image, which is used to condition the output for the generation of the subsequent HDR images. We propose a novel methodology of employing a CNN-based decoder network to enhance the obtained latent representation, yielding superior conditioning over input LDR images. The proposed loss function employed in this work integrates several terms, including reconstruction loss for the autoencoder architecture, multiscale training loss \cite{hoogeboom2023simple}, and perceptual loss \cite{zhang2018unreasonable}. Additionally, we introduce a novel loss function named "Exposure Loss," which helps in achieving optimal balance in the exposure of the reconstructed images. By prioritizing this metric, our approach significantly improves the quality of the resulting output images. Incorporating these terms in the loss function helps achieve high-quality HDR image reconstruction from a single LDR image. The effectiveness of the proposed approach is evaluated through a series of experiments, which demonstrate its superiority in terms of both reconstruction quality and convergence rate compared to existing state-of-the-art methods.

\begin{figure}[]
\center
\includegraphics[width=\columnwidth]{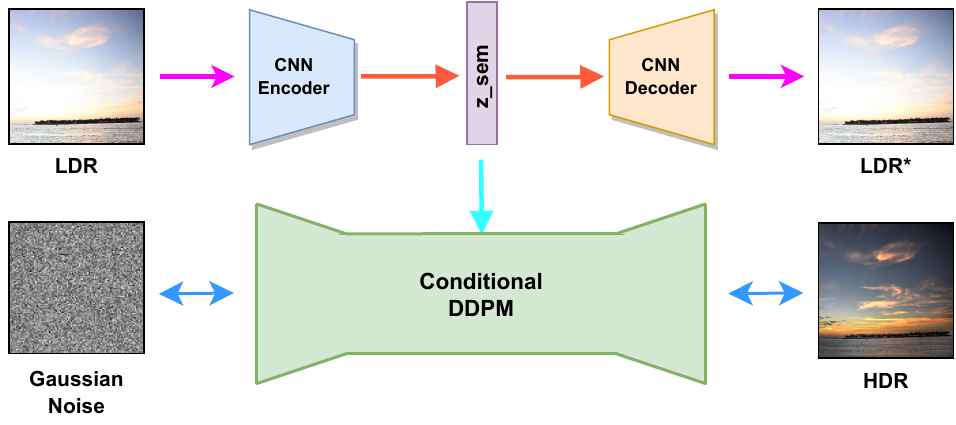}
\caption{The figure illustrates conditional diffusion architecture. In it, \textit{z\_sem} represents the latent representation of the input LDR image, and LDR$^{*}$ is the output reconstruction of the input LDR image.}
\label{fig: diff}
\vspace{-5mm}
\end{figure}
\vspace{-1mm}
\section{Background}

The family of generative models that includes score-based generative models and diffusion-based models (DPMs) has been shown effective in modeling the target distribution through a denoising process with varying noise levels. These models can accurately transform an arbitrary Gaussian noise map of the prior distribution, $\mathcal{N}(\mathbf{0}, \mathbf{I})$, into a clear image sample after multiple denoising passes. To achieve this, Ho et al. \cite{ho2020denoising} proposed a method for learning a function, $\epsilon_{\theta}(x_{t}, t)$, that predicts the corresponding noise of a noisy image, $x_t$, using a UNet architecture \cite{ronneberger2015u}. The model is trained using a loss function, $\left|\epsilon_\theta\left(x_t, t\right)-\epsilon\right|$, where $\epsilon$ is the noise added to $x_0$ producing $x_{t}$. The present formulation represents a simplified adaptation of the variational lower bound for the marginal log-likelihood, which has gained widespread adoption within the research community \cite{dhariwal2021diffusion} \cite{ho2022imagen} \cite{nichol2021improved} \cite{song2020denoising}.

\vspace{-2mm}

\subsection{Forward Diffusion}
The forward diffusion adds Gaussian noise to a given HDR image in a series of T steps. The initial image is sampled from the training data distribution $q(x)$, and a variance schedule $\beta_t \in (0,1)$ controls the noise step sizes. Specifically, at each step, the noisy version of the image $\mathbf{x_{0}}$ is generated by $q\left(\mathbf{x_t} \mid \mathbf{x_{t-1}}\right)=\mathcal{N}\left(\sqrt{1-\beta_t} \mathbf{x_{t-1}}, \beta_t \mathbf{I}\right)$ \cite{preechakul2022diffusion}, resulting in a sequence of samples $x_{1},x_{2},...x_{T}$. Due to Gaussian diffusion, we can produce the noisy version of the image $\mathbf{x_0}$ at any timestep $t$ as 

\begin{equation}
    q\left(\mathbf{x}_t \mid \mathbf{x}_0\right)=\mathcal{N}\left(\sqrt{\alpha_t} \mathbf{x}_0,\left(1-\alpha_t\right) \mathbf{I}\right) ; \alpha_t=\prod_{s=1}^{t}\left(1-\beta_s\right)
\end{equation}

In most cases, $\alpha$-cosine schedule is used among the various possible choices for selecting the variance scheduler. However, for higher resolutions images, Hoogeboom et al. \cite{hoogeboom2023simple} showed that the noise added by $\alpha-$cosine schedule is not enough. Hence, here we use a modified noise scheduler introduced by \cite{hoogeboom2023simple}. The noise schedule is adjusted to hold the signal-to-noise ratio (SNR) constant at $64 \times 64$ resolution scale.  

\vspace{-0.6cm}

\begin{equation}
    \log \mathrm{SNR}_{\text {shift } 64}^{256 \times 256}(t)=-2 \log \tan (\pi t / 2)+2 \log (64 / 256)
\end{equation}


\subsection{Reverse Diffusion}
Upon successfully adding sequential noise to an input HDR image, our focus shifts to the inverse process, namely, the distribution $p\left(\mathbf{x}_{t-1} \mid \mathbf{x}_t\right)$. We utilize our AttenRecResUNet Fig. \ref{fig: diff} to model this distribution. If the time gap between $t-1$ and $t$ is negligible (i.e., $T = \infty$), the probability function takes a Gaussian form, expressed as $\mathcal{N}\left(\mu_\theta\left(\mathbf{x}_t, t\right), \sigma_t\right)$ \cite{ho2020denoising}. Various techniques can be employed to model this distribution, including the standard epsilon loss $\epsilon_\theta\left(x_t, t\right)$ \cite{ho2020denoising}. It is vital to note that since $T=\infty$ is an unrealistic assumption in practice, DPMs can only offer approximations.

\vspace{-2mm}

\subsection{Conditional Diffusion}

Similar to other generative frameworks, diffusion models can be made to sample conditionally given some variable of interest $p_\theta\left(x_0 \mid y\right)$ like a class label or a sentence description. Particularly in our case of generation of HDR images given a single LDR input, we want our output that is generated by starting from Gaussian noise to be conditioned on the input LDR image. \cite{dhariwal2021diffusion} show that guiding the diffusion process using a separate classifier can help. In this setup, we take a pre-trained classifier to guide the reverse diffusion process. Specifically, we push process in the direction of the gradient of the target label probability. The downside of this approach is the reliance of another guiding network. An alternative approach proposed by \cite{ho2022classifier} eliminates this reliance by using special training of the diffusion model itself to guide the sampling.

\begin{equation}
\hat{\epsilon}_\theta\left(x_t, t, y\right)=\epsilon_\theta\left(x_t, t, \phi\right)+s \cdot \epsilon_\theta\left(x_t, t, y\right)-\epsilon_\theta\left(x_t, t, \phi\right)
\end{equation}

During training, the conditioning label, denoted by y, can be assigned a null label with a certain probability. At the inference stage, an artificial shift towards the conditional direction is applied to the reconstructed samples using a parameter termed the guidance scale ($s$) to distance them from the null label and thus enhance the effect of conditioning. This approach has been shown to yield superior sample quality based on human evaluation compared to classifier guidance, as reported in \cite{nichol2021glide}.

\begin{figure}[!tbh]
\center
\resizebox{0.49\textwidth}{!}{
\includegraphics[width=\columnwidth]{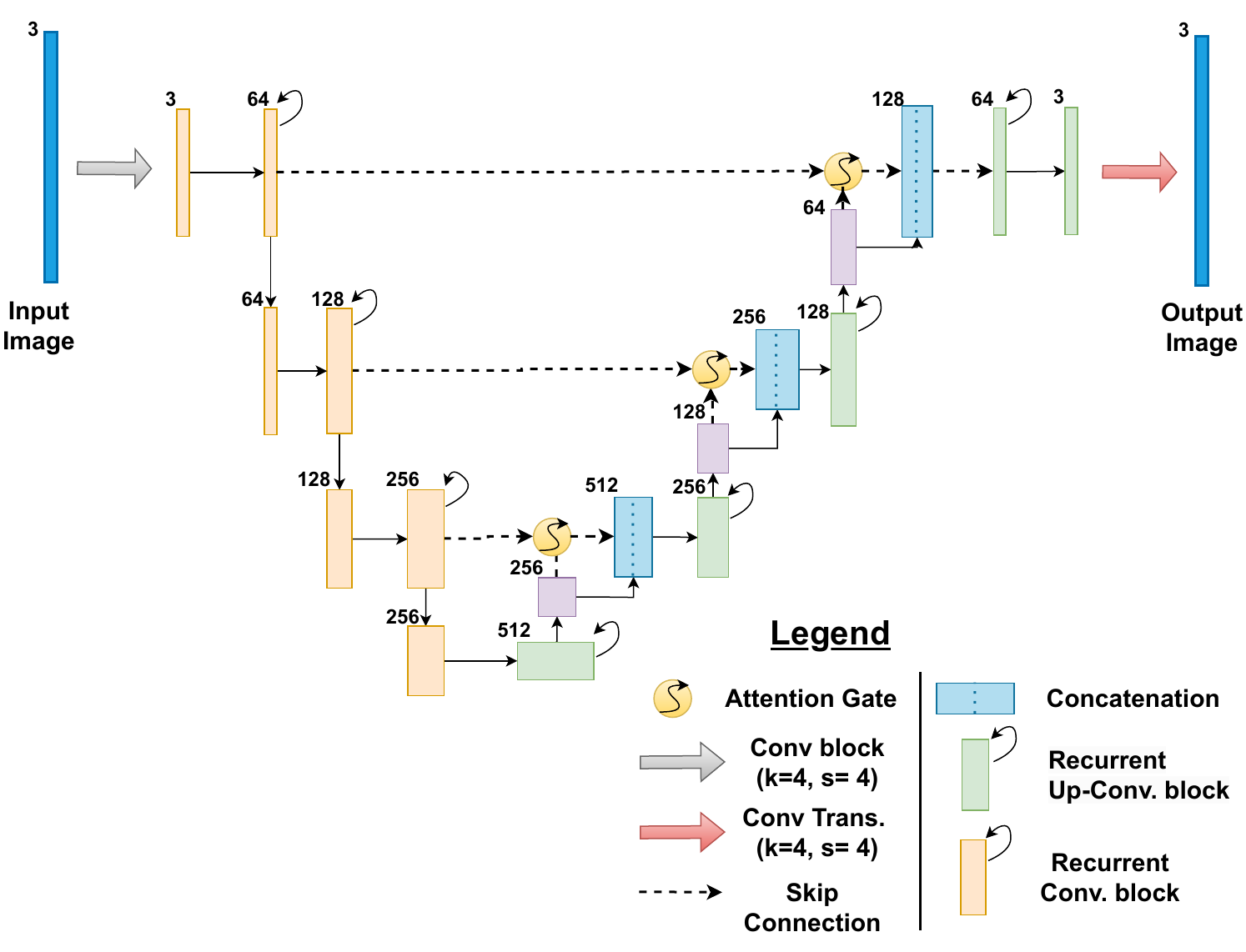}
}
\caption{Architecture of Attention Recurrent Residual UNet, used for estimating Gaussian noise in image per time-step $t$. Downsampling is applied with max-pooling = 2, and upsampling with scale = 2}
\label{fig: diff}
\vspace{-5mm}
\end{figure}

\begin{table*}[!tbh]
    \centering
    \resizebox{\textwidth}{!}{%
    \begin{tabular}{ccccccc}
        \textbf{Input LDR} &
        \textbf{DrTMO} &
        \textbf{HDRCNN} &
        \textbf{ExpandNet} &
        \textbf{SingleHDR} &
        \textbf{Ours} &
        \textbf{Ground Truth} \\
        \includegraphics[width=0.13\linewidth]{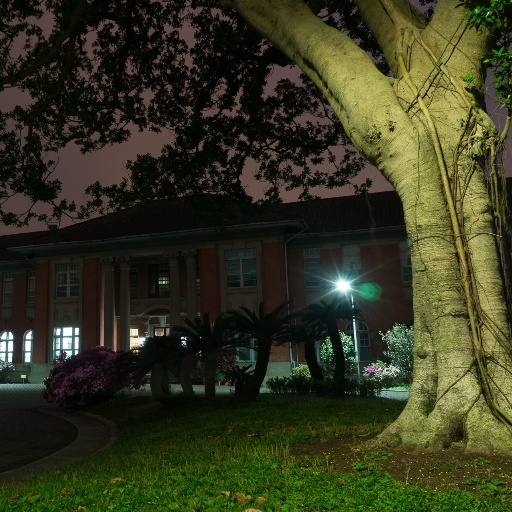} &
        \includegraphics[width=0.13\linewidth]{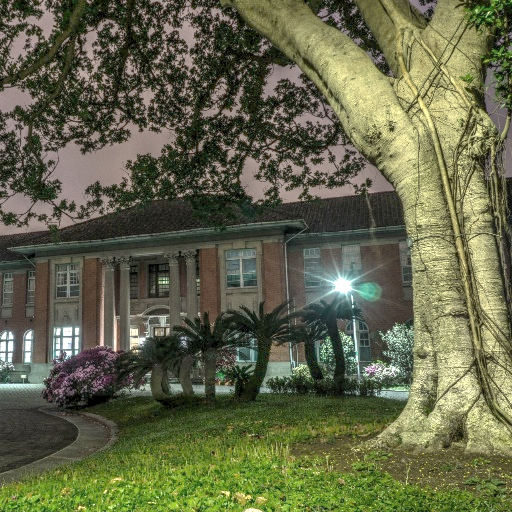} &
        \includegraphics[width=0.13\linewidth]{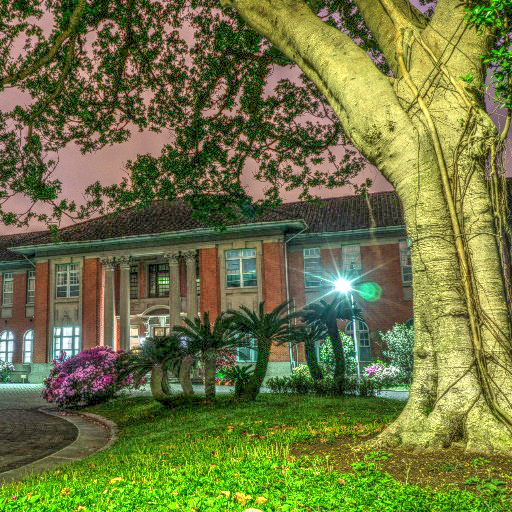} &
        \includegraphics[width=0.13\linewidth]{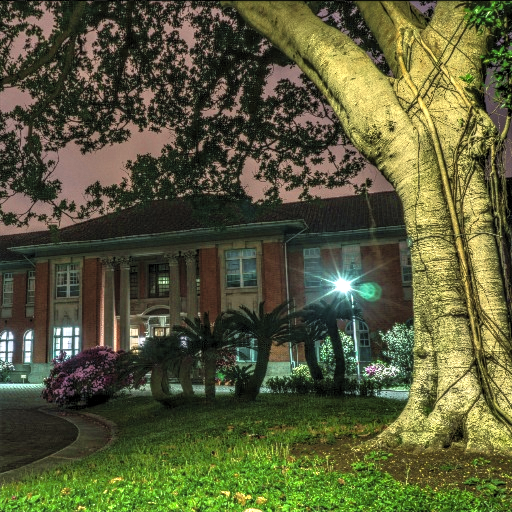} &
        \includegraphics[width=0.13\linewidth]{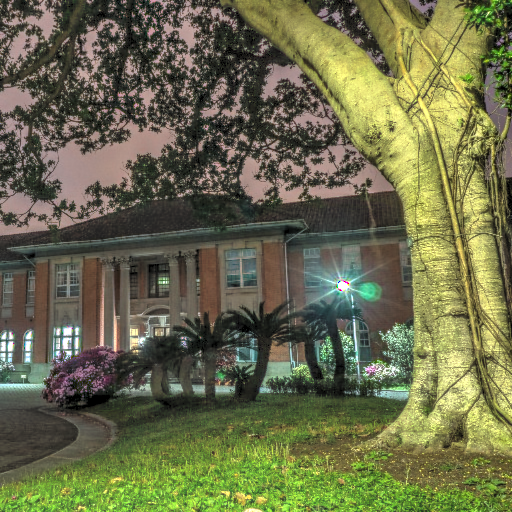} &
        \includegraphics[width=0.13\linewidth]{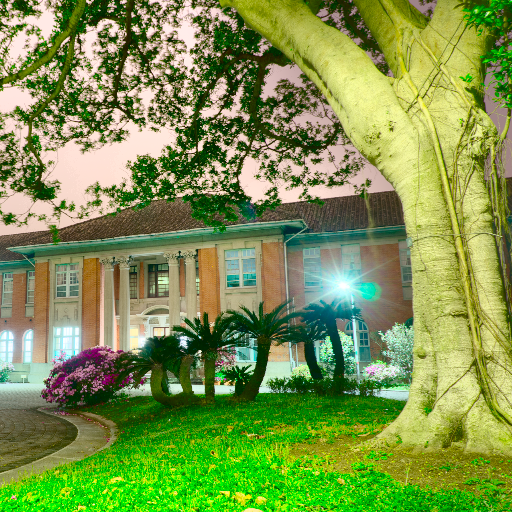} &
        \includegraphics[width=0.13\linewidth]{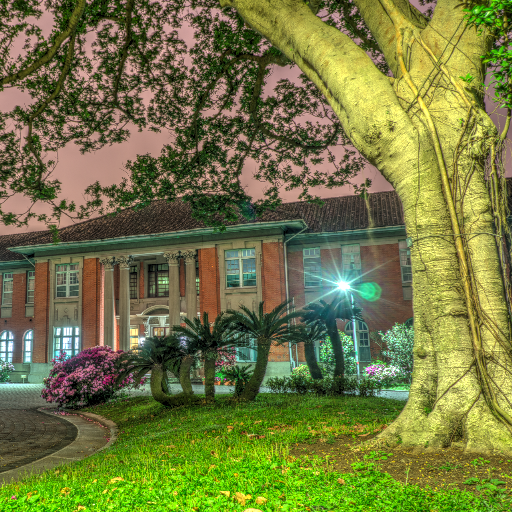} \\
        \includegraphics[width=0.13\linewidth]{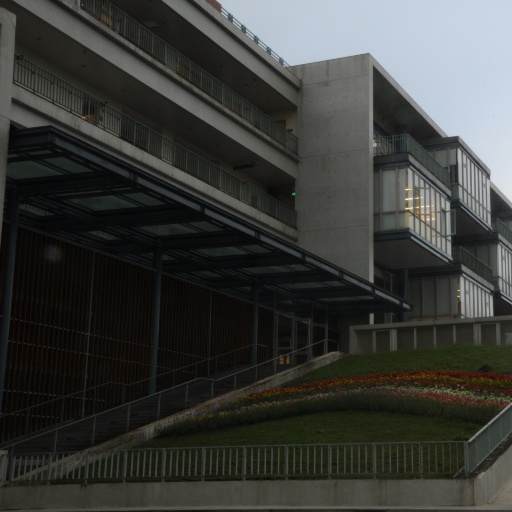} &
        \includegraphics[width=0.13\linewidth]{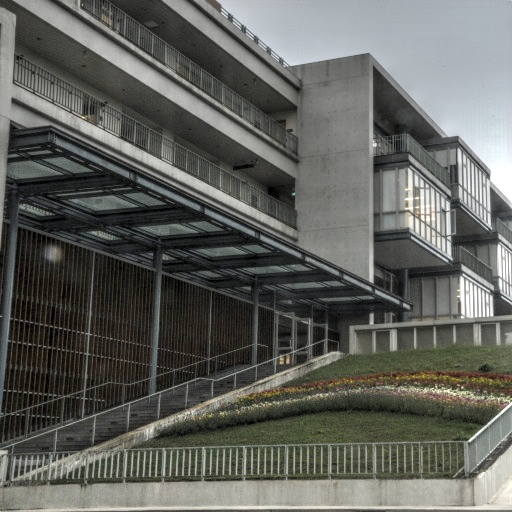} &
        \includegraphics[width=0.13\linewidth]{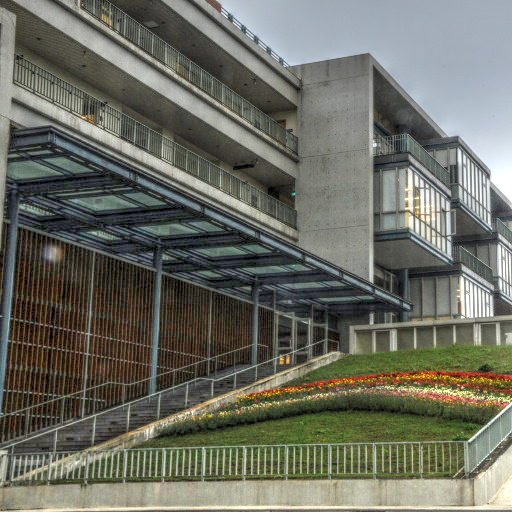} &
        \includegraphics[width=0.13\linewidth]{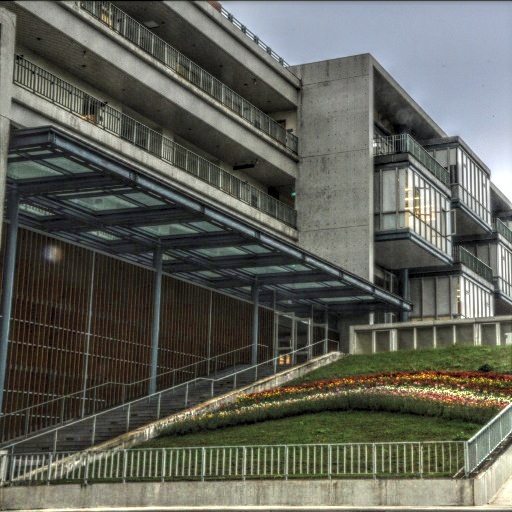} &
        \includegraphics[width=0.13\linewidth]{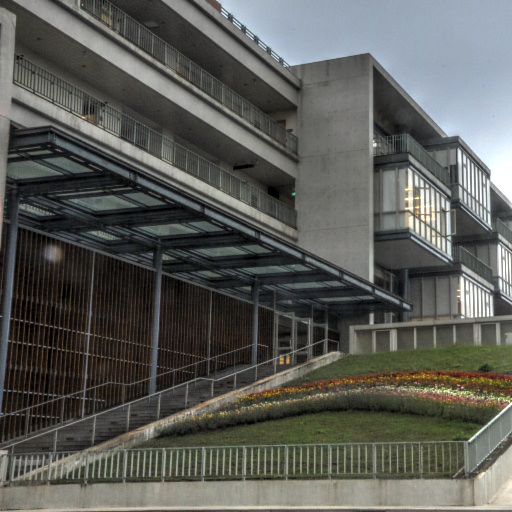} &
        \includegraphics[width=0.13\linewidth]{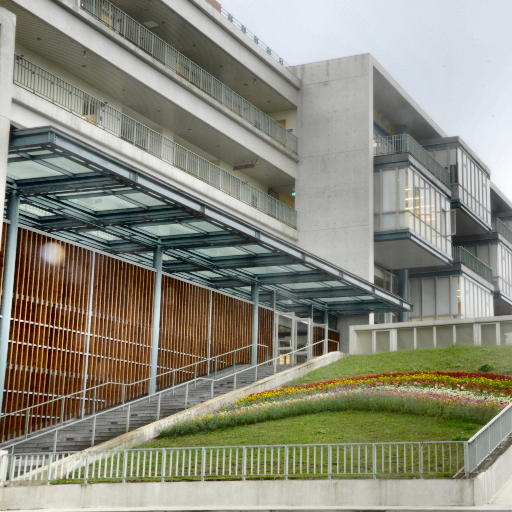} &
        \includegraphics[width=0.13\linewidth]{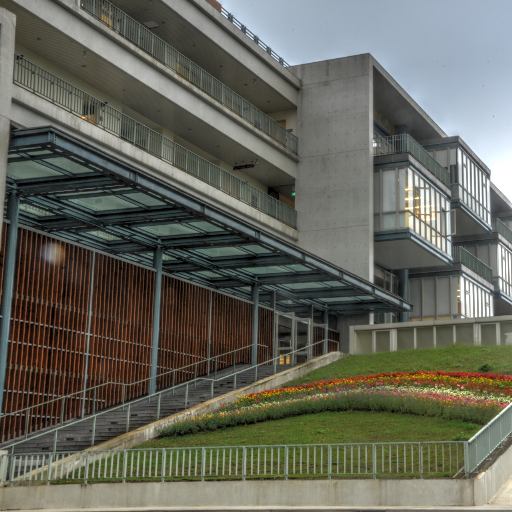} 
    \end{tabular}
    }
    \caption{Qualitative comparison of HDR image generated from our proposed approach with the previous state-of-the-art methods.}
    \label{tab:my-image-table}
    
\end{table*}

\begin{table*}[!tbh]
\centering
\begin{tabular}{lcccccc}
\toprule
                   &              & 
                   \textbf{HDR-EYE}     &              &              & \textbf{HDR-REAL}    &              \\
\hline
Method             & PSNR      & SSIM     & VDP 2.2   & PSNR      & SSIM     & VDP 2.2(1)   \\
\hline
DrTMO {\cite{endo2017deep}}      & 9.28 $\pm$ 2.98  & 0.69 $\pm$ 0.15 & 48.33 $\pm$ 5.16 & 5.54 $\pm$ 3.55  & 0.36 $\pm$ 0.26 & 43.64 $\pm$ 6.51 \\
HDRCNN {\cite{eilertsen2017hdr}}     & 16.12 $\pm$ 3.77 & 0.74 $\pm$ 0.12 & 50.75 $\pm$ 5.75 & 13.34 $\pm$ 7.68 & 0.53 $\pm$ 0.29 & 47.20 $\pm$ 7.77 \\
ExpandNet {\cite{marnerides2018expandnet}} & \textcolor{red}{\textbf{17.12 $\pm$ 4.27}} & \textcolor{blue}{\underline{0.79 $\pm$ 0.13}} & 51.09 $\pm$ 5.87 & 12.84 $\pm$ 6.90 & 0.50 $\pm$ 0.29 & 46.70 $\pm$ 7.90 \\
SingleHDR {\cite{liu2020single}} & 15.47 $\pm$ 6.65 & 0.71 $\pm$ 0.19 & \textcolor{red}{\textbf{53.15 $\pm$ 5.91}} & \textcolor{red}{\textbf{19.07 $\pm$ 7.16}} & \textcolor{blue}{\underline{0.64 $\pm$ 0.28}} & \textcolor{red}{\textbf{50.13 $\pm$ 7.74}} \\
Ours               & \textcolor{blue}{\underline{16.97 $\pm$ 4.94}} & \textcolor{red}{\textbf{0.81 $\pm$ 0.14}} & \textcolor{blue}{\underline{52.29 $\pm$ 5.82}} & \textcolor{blue}{\underline{15.76 $\pm$ 6.68}} & \textcolor{red}{\textbf{0.66 $\pm$ 0.27}} & \textcolor{blue}{\underline{49.85 $\pm$ 7.27}} \\
\bottomrule
\end{tabular}
\caption{Quantitative evaluation of our approach. The best-performing method is highlighted \textcolor{red}{\textbf{red}} and second-best  performing method is indicated by \textcolor{blue}{\underline{blue}} for each metric.}

\label{tab: mak}
\vspace{-5mm}
\end{table*}

\subsection{Diffusion model for higher resolution}

The diffusion model struggles to converge when dealing with higher resolution images  ($256 \times 256$) \cite{hoogeboom2023simple}. To overcome this limitation, we have incorporated three methodologies, (1) modified noise schedules, (2) multiscale loss function and (3) architecture scaling,  proposed in \cite{hoogeboom2023simple}. Our proposed model can handle sufficiently large under-/over exposed regions even with relatively fewer artifacts without any explicit inverse camera pipeline and is able to compete with the state-of-the-art models \cite{liu2020single} \cite{eilertsen2017hdr} \cite{vincent2008extracting} \cite{endo2017deep} in this domain with a much more streamlined architecture.

\vspace{-3mm}
\section{Methodology}

In this study, we have proposed a framework for a single image-based LDR-HDR reconstruction using a probabilistic diffusion model \cite{ho2020denoising}. The framework consists of an autoencoder and conditional DDPM, as shown in Fig.\ref{fig: diff}. The autoencoder model is used to generate encoding for the LDR image, which is later added to conditional DDPM. In the forward pass of the ddpm method, we first encode HDR images to isotropic gaussian noise, and the task of the backward pass is to reconstruct the HDR image using sampled gaussian noise by conditioning it on the encoded LDR image.  We utilize four loss functions to ensure the convergence of the framework during training.  

\vspace{-3mm}
\subsection{Multiscale training Loss}
For high-resolution images, the unweighted loss on $\epsilon$$_t$ introduced by \cite{ho2020denoising} fails to converge due to the domination of high-frequency details \cite{hoogeboom2023simple}. Hence, here we used multiscale training loss \cite{hoogeboom2023simple}, which comprises the weighted sum of losses of multiple resolutions.

\begin{equation}
    \tilde{L}_\theta^{256 \times 256}(\boldsymbol{x})=\sum_{s \in\{32,64,128, 256\}} \frac{1}{s} L_\theta^{s \times s}(\boldsymbol{x})
\end{equation}

where $L_\theta^{s \times s}$ denotes, 

\hspace{-0.51cm}
\begin{equation}
    L_{\theta}^{d \times d}(\boldsymbol{x}) = \frac{1}{d^2} \mathbb{E}_{\boldsymbol{\epsilon}, t}\|\mathrm{D}^{d \times d}[\boldsymbol{\epsilon}]-\mathrm{D}^{d \times d}[\hat{\boldsymbol{\epsilon}}_{\theta}(\alpha_{t} \boldsymbol{x}+\sigma_{t} \boldsymbol{\epsilon}, t)]\|_{2}^{2}
\end{equation}


and $D^{d \times d}$ downsamples to $d \times d$ resolution. Here $d$ will take values belonging to $\{32,64,128,256\}$




\subsection{Reconstruction Loss}
The decoder network is trained using reconstruction loss, computed on the image generated by the decoder using the compact latent space derived from the low dynamic range (LDR) image. This loss is formally defined as follows:
\begin{equation}
L_{rec} =   \parallel  \xi_{LDR}  -  \tilde{\xi}     \parallel^{2} 
\end{equation}

Here, $\xi_{LDR}$ represents the ground truth LDR image, and $\tilde{\xi}$ denotes the decoded output of the autoencoder network. The reconstruction loss is designed to ensure that the autoencoder creates a more meaningful semantic representation of the input LDR image, which is then used to condition our AttenRecResUNet (Fig. \ref{fig: diff}) architecture in the reverse diffusion process.

\subsection{Perceptual Loss}
The Learned Perceptual Image Patch Similarity (LPIPS) \cite{zhang2018unreasonable} metric is utilized to assess the perceptual similarity of two images, and has been demonstrated to align with human perception. 
Here, the LPIPS score is employed to expedite model convergence by ensuring that the higher-level semantics of the predicted noisy image align perceptually with those of the original noisy image at each time step.

\subsection{Exposure Loss}
In most cases, we want the exposure of the output HDR image to be inverse with respect to the input LDR image, i.e., if an LDR image is over-exposed, we would want to push the gradients of our model in the opposite direction, making sure that the output HDR image has balanced exposures. To be able to achieve this, we introduce a new loss defined as Exposure loss $L_{exp}$ that helps in achieving the opposite gradient flow. 

\vspace{-11pt}

\begin{equation}
    {L}_{exp}= -\zeta *\left(\left|\frac{\sum_{p} x_t}{\sum_{p} max(x_t,1)}- \frac{\sum_{p} x_{ldr}}{\sum_{p} max(x_{ldr},1)} \right|\right)
\end{equation}


Here $\sum_{p}$ denotes pixel-wise summation, $x_t$ denotes the normalized predicted HDR image, and $x_{ldr}$ denotes the normalized input LDR image. The negation ensures that we minimize the penalty for the loss when there is a contrast between the exposures of the input LDR and the predicted HDR image. Scaling factor $\zeta$ was introduced  to balance the impact(magnitude) of $L_{exp}$ with the other terms, ensuring a fair contribution from each loss term during optimization. The final loss function that we use to train the proposed frames is as follows:

\vspace{-11pt}

\begin{equation}
    \mathcal{L}_{T} = L_{MSTL} + L_{rec} + L_{lpips} + L_{exp}
\end{equation}

\section{Experiments and Results}

We evaluate our proposed method by conducting a comparison against several single-image-based HDR reconstruction approaches, including DrTMO \cite{endo2017deep}, HDRCNN \cite{eilertsen2017hdr}, ExpandNet \cite{marnerides2018expandnet}, and SingleHDR \cite{liu2020single}. We performed a comparison on two publicly available datasets HDR-Eye and HDR-REAL (test split). There are 1838 LDR-HDR image pairs in the HDR-REAL dataset test set and 46 LDR-HDR image pairs in HDR-Eye. The direct inference was conducted on publicly available pre-trained models of these approaches to obtain output HDR images.

The qualitative comparison is illustrated in Table \ref{tab:my-image-table}, and for the quantitative evaluation, we used three different metrics: HDR-VDP 2.2 \cite{mantiuk2011hdr}, PSNR, SSIM \cite{wang2004image} metrics. Table \ref{tab: mak} shows the scores obtained on HDR-Eye and HDR-Real Dataset (Test split) for the three metrics. Table \ref{tab:my-image-table} \& \ref{tab: mak} showcases that our model performs commendably when compared to the state-of-the-art models.


\section{Ablations}


\subsection{Autoencoder Diffusion}
We introduce a novel deep CNN based decoder network that focuses on enhancing the quality of the latent representation formed from our encoder network that is used in the conditioning of our AttenRecResUNet (Fig. \ref{fig: diff}) during the reverse diffusion process. We validate the results of our Autoencoder Diffusion Model with the following variants:

\begin{itemize}
    \item \textbf{Autoencoder Diffusion} - This is the complete autoencoder architecture used as a subpart of our proposed network.
    \item \textbf{Encoder Diffusion} - In this model, we remove the Decoder module from our autoencoder Diffusion architecture.
\end{itemize}

 Table \ref{tab:bhosda} \& \ref{tab:confusion-matrix}  qualitatively and quantitatively shows that adding the decoder block in our model causes the output HDR image to become significantly sharper compared to its variant.


\subsection{Loss function}
We also performed ablation studies on the importance of exposure loss. In Case 1 (Exposure P.), we included $L_{exp}$ in the total loss computation, which was defined as total Loss = $L_{MSTL}$ + $L_{rec}$ + $L_{lpips}$ + $L_{exp}$. In Case 2 (Exposure A.), we excluded $L_{exp}$ and kept all other loss components intact. Table \ref{tab:bhosda} \& \ref{tab:confusion-matrix}  qualitatively and quantitatively demonstrates that the addition of the Exposure Loss improves the exposure balance of the output HDR as compared to its variant.


\begin{table}[!tbh]
\resizebox{0.49\textwidth}{!}{%

\begin{tabular}{cccc}
\toprule
\textbf{Method}       & \textbf{PSNR}      & \textbf{SSIM}     & \textbf{VDP 2.2}   \\
\hline
 \begin{tabular}[c]{@{}c@{}}Vanilla Conditional  \\Diffusion\end{tabular} & 16.36 $\pm$ 4.55 & 0.74 $\pm$ 0.11 & 50.93 $\pm$ 5.68 \\
\hline
\begin{tabular}[c]{@{}c@{}}Autoencoder \\Diffusion\end{tabular}  & 16.71 $\pm$ 4.69 & 0.79 $\pm$ 0.19 & 51.67 $\pm$ 5.79 \\

\hline

\begin{tabular}[c]{@{}c@{}} Autoencoder \\ Diffusion + Total Loss\end{tabular} & 16.97 $\pm$ 4.94  & 0.81 $\pm$ 0.14 & 52.29 $\pm$ 5.82
\\
\bottomrule
\end{tabular}
}

\caption{Evaluation of our approach on HDR-Eye dataset}
\label{tab:bhosda}
\vspace{-3mm}
\end{table}

 \begin{table}[!htbp]
\centering
\begin{tabular}{c c c}
\toprule
 & \textbf{Exposure A.} &  \textbf{Exposure P.} \\
\rotatebox[origin=l]{90}{\textbf{ 
  Decoder A.}} & \includegraphics[width=0.3\linewidth]{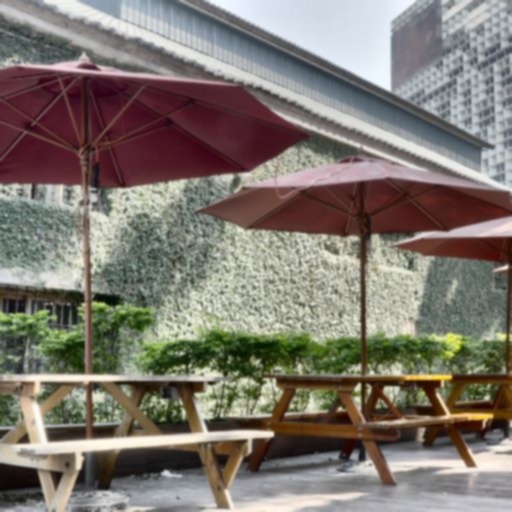} & \includegraphics[width=0.3\linewidth]{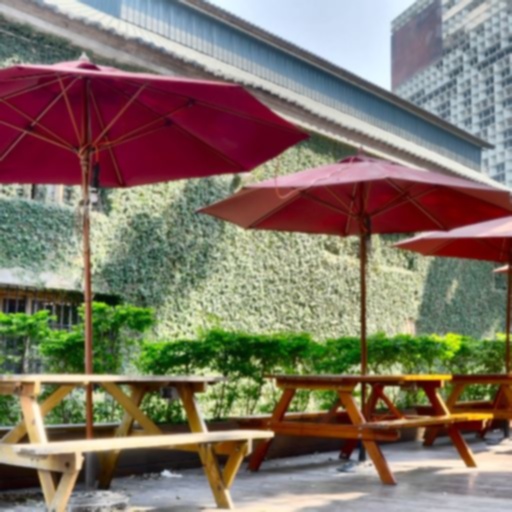} \\
 \rotatebox[origin=l]{90}{\textbf{ 
Decoder P.}} & \includegraphics[width=0.3\linewidth]{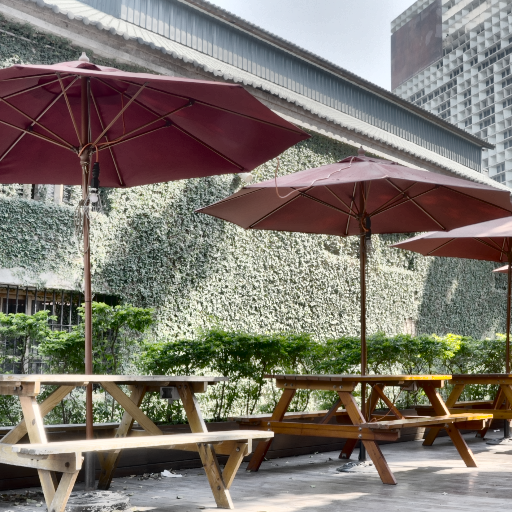} & \includegraphics[width=0.3\linewidth]{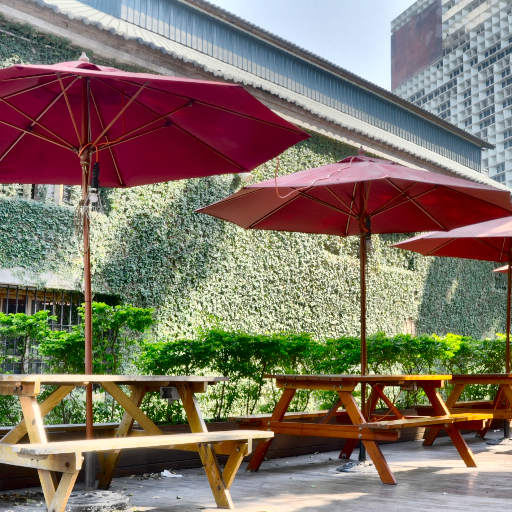} \\
\bottomrule
\end{tabular}

\caption{Images plotted in the form of a confusion matrix to demonstrate the improvement of output on the addition of decoder block and Exposure loss. Here A. means absent, and P. means present.}
\label{tab:confusion-matrix}
\vspace{-5mm}
\end{table}

\vspace{-5pt}

\section{Limitation and Future Works}
With Exposure loss and decoder network in our architecture, we were able to substantially increase the speed of convergence as well as improve the quality of the generated HDR samples. However, in areas with lesser artifacts in the saturated regions, the quality of the output HDR images can be further improved by incorporating an inverse camera pipeline as done by \cite{liu2020single} and \cite{raipurkar2021hdr}. By incorporating masking on under and over-exposed regions to treat them separately, the quality of the model in more stringent conditions can be improved.

\vspace{-5mm}
\section{Conclusion}
The proposed framework reconstruct an HDR image from a single LDR image and is based on conditional diffusion with an autoencoder. We show that adding noise in HDR images during the forward pass of diffusion can help the network reconstruct lost information in LDR images during the backward pass. We also proposed a novel loss, named Exposure loss, that focused on explicitly directing the gradients in the direction opposite to the saturation, thus enhancing the quality of the results. Additionally, to ensure faster convergence and learn the semantics of the artifacts in the saturated regions quicker, we used perceptual loss. We show the effectiveness of the proposed framework over the previous state-of-the-art methods through multiple experiments and ablation studies.

\bibliographystyle{IEEEbib}
\bibliography{custom}

\end{document}